\documentclass[letterpaper, 10 pt, conference]{ieeeconf}  

\IEEEoverridecommandlockouts                              

\usepackage{graphicx} 
\usepackage{amsmath} 
\newcommand\norm[1]{\left\lVert#1\right\rVert}
\usepackage{amssymb}  
\usepackage{todonotes}
\usepackage{nccmath}	
\usepackage{amsbsy}
\usepackage{mathtools}
\usetikzlibrary{automata, positioning}
\usetikzlibrary{fit,positioning,arrows,automata}
\usepackage{multirow}
\usetikzlibrary{shadows,positioning}
\usepackage{cite}
\usepackage{subcaption}
\usepackage{booktabs}
\usepackage{tikz}
\usetikzlibrary{external}
\tikzset{external/mode=graphics if exists}
\usepackage{pgfplots}

\pgfplotsset{grid style={dotted, gray}}
\pgfplotsset{minor grid style={dotted,gray}}
\pgfplotsset{every tick label/.append style={font=\tiny}}
\pgfplotsset{every axis/.append style={font=\small}}
\pgfplotsset{ylabel near ticks}
\pgfplotsset{xlabel near ticks}
\newlength\figureheight 
\newlength\figurewidth

\let\oldmissingfigure\missingfigure
\renewcommand{\missingfigure}[2][]{\tikzexternaldisable\oldmissingfigure[#1]{#2}\tikzexternalenable}

\newcommand{\markku}[1]{}
\newcommand{\mattia}[1]{}
\newcommand{\joni}[1]{}
\newcommand{\alberto}[1]{}

\title{\LARGE \bf
	Segmenting and Sequencing of Compliant Motions
}

\author{Tesfamichael Marikos Hagos$^1$, Markku Suomalainen$^1$ and Ville Kyrki$^1$
\thanks{*This work was supported by Academy of Finland, decision 286580.}
\thanks{T. M. Hagos, M. Suomalainen and V. Kyrki are with School of Electrical Engineering, Aalto University, Finland {\tt\small tesfam19@yahoo.com, markku.suomalainen@aalto.fi, ville.kyrki@aalto.fi}}%
}

\begin{document}

\maketitle
\thispagestyle{empty}
\pagestyle{empty}

\begin{abstract}

This paper proposes an approach for segmenting a task consisting of compliant motions into phases, learning a primitive for each segmented phase of the task, and reproducing the task by sequencing primitives online based on the learned model. As compliant motions can ``probe'' the environment, using the interaction between the robot and the environment to detect phase transitions can make the transitions less prone to positional errors. This intuition leads us to model a task with a non-homogeneous Hidden Markov Model (HMM), wherein hidden phase transition probabilities depend on the interaction with the environment (wrench measured by an F/T sensor). Expectation-maximization algorithm is employed in estimating the parameters of the HMM model. During reproduction, the phase changes of a task are detected online using the forward algorithm, with the parameters learned from demonstrations. Cartesian impedance controller parameters are learned from the demonstrations to reproduce each phase of the task. The proposed approach is studied with a KUKA LWR4+ arm in two setups. Experiments show that the method can successfully segment and reproduce a task consisting of compliant motions with one or more demonstrations, even when demonstrations do not have the same starting position and external forces occur from different directions. Finally, we demonstrate that the method can also handle rotational motions. 

\end{abstract}

\section{INTRODUCTION}
Position uncertainty is a major difficulty in manipulation tasks such as assembly. Position control in such cases can cause high contact wrenches (force and torque), ultimately leading to behaviors such as breakage or unstable control \cite{de1988compliant}. Force control can be used to mitigate the adverse effects by controlling the interaction forces with the environment while moving in contact. The motions requiring interaction with the environment are called compliant motions. 

\begin{figure}[htb]
	\centering
	\includegraphics[width=9cm]{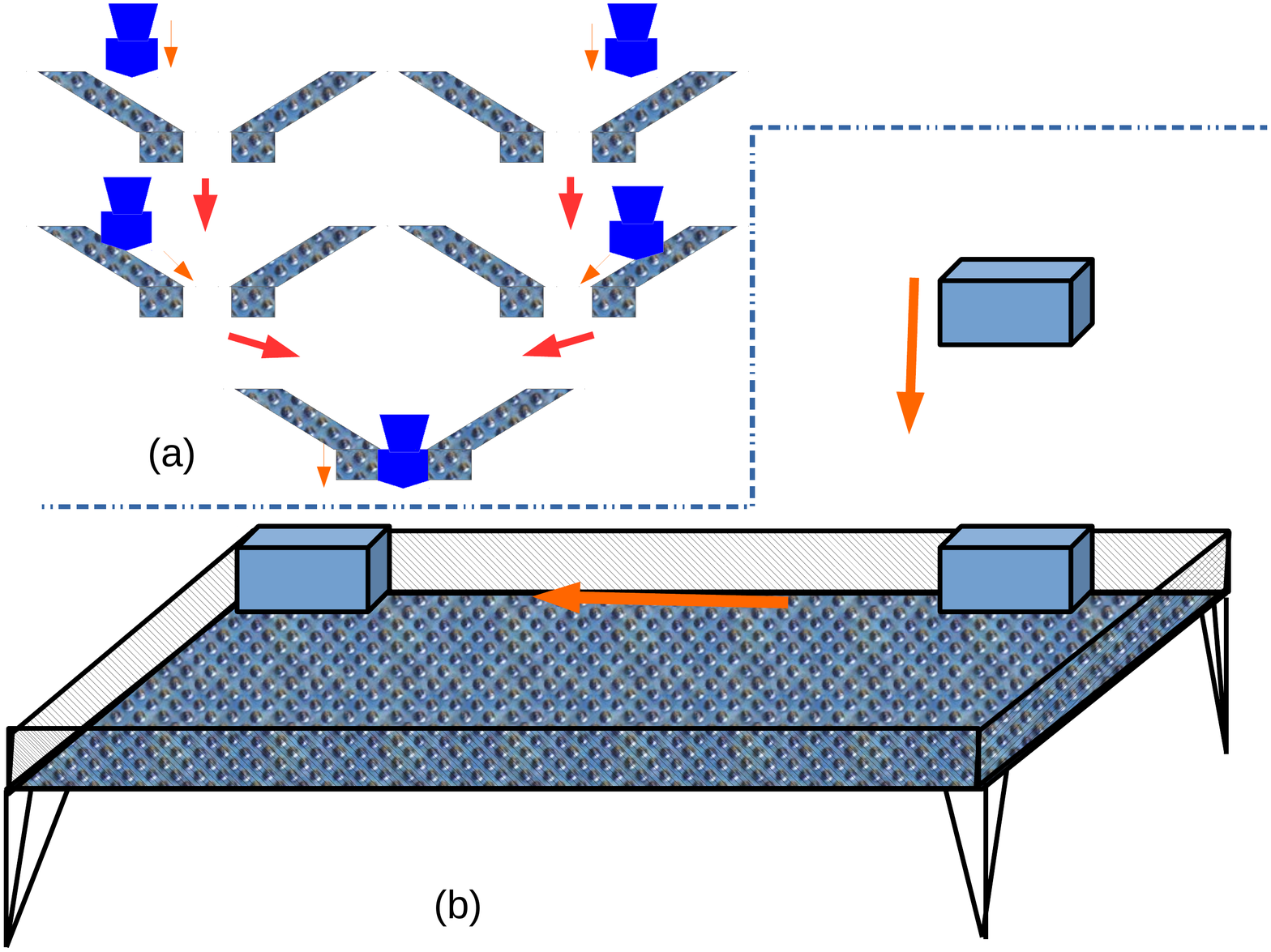}
	\caption{Compliant motion used (a) to align workpieces and (b) to place a box at the corner of the table}
\vspace{-2em}
\label{setup}
\end{figure}

Besides direct force control, another approach for performing compliant motions is impedance control. An impedance controller uses a force-displacement model between the end-effector and environment, modelled as a linear spring with stiffness along an axis of interest \cite{part1985impedance}. Such a controller can be used for both non-contact 
and contact phases of a task. However, transferring a skill requiring impedance control to a robot is complicated and planning such motions automatically is computationally infeasible. For that reason, the study of learning mechanisms for compliant motions has received significant interest.

Learning from demonstration (LfD) is an approach in which a robot can learn compliant motions from human demonstrations \cite{cabras2010contact}, which facilitates the transfer of skills to a robot.
LfD can be combined with compliant motions by taking advantage of the structure of the environment to reduce the need for positional accuracy \cite{suomalainen2017, suomalainen2016learning}, for example to align workpieces (Fig.~\ref{setup}a) or to place an object in a corner (Fig.~\ref{setup}b). However, the recent work \cite{suomalainen2017,suomalainen2016learning} can learn only individual motions, being unable to learn and reproduce a task consisting of a sequence of motions such as Fig.~\ref{setup}b. The segmentation of human motions into compliant phases\footnote{Can be referred to as mode in hybrid dynamics} is non-trivial since the actual motion direction during a phases may vary due to differences in interaction forces. For example in Fig.~\ref{setup}a sliding along either of the chamfers must be detected as the same phase. 

This paper expands the idea of linear compliant motions by proposing an approach that segments the task into phases consisting of a single motion, learns the controller parameters for each phase using the method presented in \cite{suomalainen2017}, and detects the phase transitions during reproduction, therefore learning to reproduce entire demonstrated tasks after one or more demonstrations.




We propose a method that allows segmenting a human demonstration of a task  into phases. We explore how to use HMM probabilistic encoding approaches for modelling observations and state transition uncertainties of a manipulation task as used in ~\cite{kroemer2014learning,kroemer2015towards}. We analyse how the model order can be predicted and how the model parameters are learned from multiple demonstrations. We model the phases of a task as hidden discrete variables. The current phase depends on the previous phase and the current measured interaction with the environment. Thus the phase change occurs when the measured interactions with environment change. The learned model can be used for sequencing compliant motion primitives from \cite{suomalainen2017, suomalainen2016learning} to reproduce a complete task.   


\section{RELATED WORK}\label{RW}

There has been considerable work on segmenting a manipulation task into subtasks based on interaction measurement. 
Some of the works use model-based approaches and require geometrical knowledge of the task~\cite{hirai1993kinematics, bicchi1993contact, farahat1995identifying, bruyninckx1995kinematic}. Such an approach can be discrete, estimating the the contact formation only, as Hovland et al.~\cite{hovland1998hidden}, who use HMM for monitoring contact formations during an assembly task. Another approach is the hybrid approach, which estimates the continuous geometric properties as well, given the state graph of the possible contact formations~\cite{meeussen2007contact, gadeyne2005bayesian}. However, approaches that exploit the geometric knowledge of the manipulated object require the analysis of geometrical and dynamical constraint equations, which is often a tedious task. 

Other works on segmentation avoid using a model of the object. Skubic and Volz~\cite{skubic2000identifying} use a probabilistic classifier for contact state identification while measuring confidence level from force signals. Jasim et al.~\cite{jasim2013ts} develop a fuzzy inference system that can model possible contact states using the available poses and interaction wrenches. In their further work, Jasim et al.~\cite{jasim2014contact} also propose a method that uses Gaussian mixture model for contact state classification. Cabras et al.~\cite{cabras2010contact} use a supervised learning algorithm to perform segmentation based on the minimization of a nonconvex function which may result in local minima trapping. Even though the above data-driven approaches do not depend on geometrical knowledge, they cannot identify motion direction for tasks consisting of compliant motions.

In the context of LfD, stochastic approaches such as HMMs facilitate the generalization of a task without knowing the geometric parameters of the object. Niekum et al. \cite{niekum2012learning} show how to use Beta-Process autoregressive HMM for segmenting a task. Relevant work by Kroemer et al.~\cite{kroemer2014learning,kroemer2015towards} tries to perform offline segmentation of a demonstrated task. In segmenting in-contact tasks such as pushing a box, Kroemer et al.~\cite{kroemer2014learning} propose an HMM based method in which the phase transition depends on position change. In their further work, Kroemer et al.~\cite{kroemer2015towards} propose another model which models transitions between phases as events, allowing the algorithm to model the entry and exit conditions of phases.
However, for compliant motions where the environment is altering the motion and the robot is effectively probing the environment, the interaction with the environment could present a more coherent cue for detecting the phase change than the position, while also reducing the need for positional accuracy.
Our method exploits the wrench measurements to detect the time-varying phase transition instead of using a relative position.
In addition, our model manages estimation of parameters from multiple demonstrations with different starting positions, such as in Fig~\ref{setup}a. During reproduction, our method sequences Cartesian impedance controllers proposed by Suomalainen and Kyrki \cite{suomalainen2017} with parameters learned from the demonstration to complete the task.

\section{METHOD}
\label{METHOD}

The main idea is to learn a task defined by a sequence of compliant motions with linear dynamics from a human demonstration. First the human teacher performs an assembly task while the position and force data are recorded. From the recorded data we first learn the number of phases with Bayesian Information Criterion (BIC). Then our algorithm segments the whole demonstrated task (one or more demonstrations) into an according number of phases, and the impedance controller parameters for each phase are learned individually with the method presented in \cite{suomalainen2017}. During reproduction, our algorithm detects the changes in the contact dynamics and switches to the next learned impedance controller accordingly.

We use pose and wrench data recorded during kinesthetic teaching to learn the probabilistic model of a task. In the model, the phases of the manipulation task are modelled as discrete hidden variables as in \cite{kroemer2014learning}. In Sect.~\ref{Model}, we introduce the structure ofthe model and the associated notations. In Sect.~\ref{MS} we explain the model selection and in Sect.~\ref{ML} we describe how the model parameters are learned.

\subsection{Model}\label{Model}
\subsubsection{Single phase dynamics}
The task is reproduced with a sequence of impedance controllers, defined as
\begin{equation}
  \pmb{F} = K(\pmb{x}^*-\pmb{x})+C\pmb{v}+\pmb{f_{dyn}},
  \label{eqt:imp_control}
\end{equation}
where $\pmb{x}^*$ is the desired position, $\pmb{x}$ the current position, $K$ the stiffness matrix, $C\pmb{v}$ a linear damping term, and $\pmb{f_{dyn}}$ the feed-forward dynamics of the robot including gravity. The trajectory for each phase is calculated in a feed-forward manner

\begin{equation}
  \pmb{x}^*_t= \pmb{x}^*_{t-1}+\pmb{\hat{v}_{d}^*} v  \Delta t
  \label{eqt:desireddir}
\end{equation}
where $\pmb{\hat{v}_{d}^*}$ is the desired direction of motion, $v$ the velocity (speed) and $\Delta t$ the sample time of the control loop. If the parameters $\pmb{\hat{v}_{d}^*}$ and $K$ are learned properly, this model can take advantage of chamfers in assembly as depicted in Fig. \ref{setup}a and is robust towards positional uncertainty and small modifications in the environment, as shown in \cite{suomalainen2016learning}.

For the purpose of segmenting a human demonstration consisting of multiple phases, we model the state dynamics of a single phase by a linear Gaussian model, in which the next state depends on current state, measured interaction, and current phase $p(\mathbf{s}_{t+1}|\mathbf{s}_t, \mathbf{a}_t, \pmb{\rho}_t)$. The state $\mathbf{s}_t\in \mathbb{R}^{m}$ represents the position of the robot TCP. $\mathbf{a}_t\in \mathbb{R}^{d}$ is the measured interaction (contact) wrench, concatenated with 1. The distribution of the next state is then
\begin{equation}
\mathbf{s}_{t+1}\sim \mathcal{N}\left(A_{\pmb{\rho}_t}\mathbf{s}_t+B_{\pmb{\rho}_t}\mathbf{a}_t, \Sigma_{\pmb{\rho}_t}\right)
\label{eqt:statedynamics}
\end{equation}
where $A_{\pmb{\rho}_t}\in \mathbb{R}^{m\times m}$ represents the uncontrolled dynamics, $B_{\pmb{\rho}_t}\in \mathbb{R}^{m\times d}$ models compliance through interaction forces and constant offset velocity through the concatenated 1, and $\Sigma_{\pmb{\rho}_t}\in \mathbb{R}^{m\times m}$ is the covariance matrix corresponding to phase ${\pmb{\rho}_t}$.
Thus, the model assumes linear system dynamics and $B_{\pmb{\rho}_t}$ is used to model contact interaction effects and constant desired direction of motion. 

\subsubsection{Sequence of phases}
\begin{figure}[h!]
	\begin{center}
		
		\begin{tikzpicture}[scale=0.50]
		
		\tikzset{node style/.style={state, minimum width=.95cm, line width=1mm, fill=gray!20!red}}
		\node[draw=none] at (-2,0) (S0)	{};
		\node[node style] at (0, 0)	(S1)	{$\mathbf{s}_{t-1}$};
		\node[node style] at (3, 0) (S2)    {$\mathbf{s}_t$};
		\node[node style] at (6, 0)	(S3)	{$\mathbf{s}_{t+1}$};
		\node[node style] at (9, 0)(S4)		{$\mathbf{s}_{t+2}$};
		
		\tikzset{node style/.style={state, minimum width=.95cm, line width=1mm, fill=gray!20!white}}
		\node[node style] at (-2.5,-3) (P0)	{$\pmb{\rho}_{t-2}$};
		\node[node style] at (0.5, -3)	(P1)	{$\pmb{\rho}_{t-1}$};
		\node[node style] at (3.5, -3) (P2)    {$\pmb{\rho}_t$};
		\node[node style] at (6.5, -3)	(P3)	{$\pmb{\rho}_{t+1}$};
		\node[draw=none] at (9.5, -3)(P4)	{};
		
		\tikzset{node style/.style={state, minimum width=.95cm, line width=1mm, fill=gray!20!red}}
		\node[node style] at (-3,-6) (A0)	{$\mathbf{a}_{t-2}$};
		\node[node style] at (0, -6)	(A1)	{$\mathbf{a}_{t-1}$};
		\node[node style] at (3, -6) (A2)    {$\mathbf{a}_t$};
		\node[node style] at (6, -6)	(A3)	{$\mathbf{a}_{t+1}$};
		\node[draw=none] at (9, -6)(A4)	{};
		\draw[every loop, auto=right, line width=0.5mm, >=latex, draw=black, fill=black]
		(S0)    edge[right=20]  node {} (S1)      
		(S1)	edge[right=20]	node {} (S2)
		(S2)	edge[right=20]	node {} (S3)
		(S3)	edge[right=20]	node {} (S4)
		(P0)    edge[right=20]  node {} (P1)
		(P1)	edge[right=20]	node {} (P2)
		(P2)	edge[right=20]	node {} (P3)
		(P3)	edge[right=20]	node {} (P4)
		(A0)	edge[right=20]	node {} (S1)
		(A1)	edge[right=20]	node {} (S2)
		(A2)	edge[right=20]	node {} (S3)
		(A3)	edge[right=20]	node {} (S4)
		(A0)	edge[right=20]	node {} (P0)
		(A1)	edge[right=20]	node {} (P1)
		(A2)	edge[right=20]	node {} (P2)
		(A3)	edge[right=20]	node {} (P3)
		(P0)	edge[right=20]  node {} (S1)
		(P1)	edge[right=20]	node {} (S2)
		(P2)	edge[right=20]	node {} (S3)
		(P3)	edge	node {} (S4);
		\end{tikzpicture}
		
	\end{center}
	\caption{Observed interaction-based transitions auto-regressive HMM model. The red circles represent the observed variables i.e. the states $\mathbf{s}_t$ and the interaction $\mathbf{a}_t$, whereas the white circles represent the hidden phases $\pmb{\rho}_t$. }
	\label{hmm03}
\end{figure}
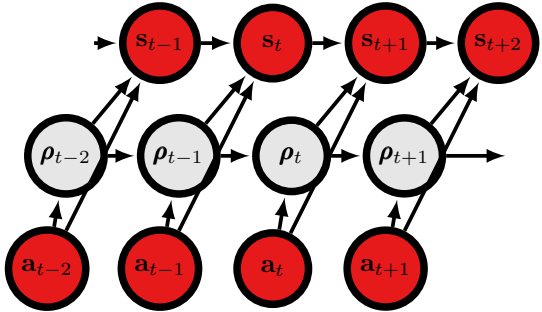

The general idea of the sequencing method was inspired by \cite{kroemer2014learning} and adapted for the compliant motion model described in the previous section. Each phase $\rho_t\in \mathbb{N}$ of a manipulation task is modelled as a hidden state of the HMM. By combining the autoregressive model from the previous section with an HMM, we end up with a graphical model given in Fig.~\ref{hmm03}. 

We define the state transition probabilities as depending on the interaction variable $\mathbf{a}_t$, since in our case the phase changes are indicated by interaction forces rather than by position. This gives phase changes robustness against position uncertainty. More precisely, the phase transition probability $p(\pmb{\rho}_t|\mathbf{a}_t, \pmb{\rho}_{t-1})$  is defined by multi-class logistic regression
\begin{equation}
p(\pmb{\rho}_t=j|\mathbf{a}_t, \pmb{\rho}_{t-1}=i) = \frac{\exp(w_{ij}\phi(\mathbf{a}_t))}{\sum_k\exp(w_{ik}\phi(\mathbf{a}_t))}
\end{equation}
where $w_{ij}\in \mathbb{R}^{d+1}$ is a weight matrix containing weights for transitioning from $\pmb{\rho}=i$ to $\pmb{\rho}=j$, and $\phi(\mathbf{a}_t)$ is an arbitrary feature function of $\mathbf{a}_t$. The first phase depends on the first interaction only, i.e. $p(\pmb{\rho}_1|\mathbf{a}_1)$, and the initial phase distribution is given by 
\begin{equation}
p(\pmb{\rho}_1|\mathbf{a}_1) = \frac{\exp(w_{0j}\phi(\mathbf{a}_1))}{\sum_k\exp(w_{0k}\phi(\mathbf{a}_1))}
\end{equation}
where $w_{0j}$ is the weight matrix for each phase. The dependency of the current phase on the previous one is important since such a model can represent hysteresis effects and transient state information by taking advantage of the wrench data. In this paper, identity was found to be sufficient as the feature function, $\phi(\mathbf{a}_t)=\mathbf{a}_t$.


Given the definition of the variables in the model, the joint probability for $T$ samples of the observed variables $\left( \mathbf{s}_{1:T}=\{\mathbf{s}_1,\cdots,\mathbf{s}_T\}, \mathbf{a}_{1:T}=\{\mathbf{a}_1,\cdots,\mathbf{a}_T\} \right)$ and the hidden variable ($\pmb{\rho}_{1:T}=\{\pmb{\rho}_1,\cdots,\pmb{\rho}_T\}$) is given by 

\begin{fleqn}
	\begin{align}
	p(\mathbf{s}_{1:T+1},\mathbf{a}_{1:T},\pmb{\rho}_{1:T})= 
	\end{align}
\end{fleqn}

\begin{equation}
p(\mathbf{s}_1)p(\pmb{\rho}_1|\mathbf{a}_1)\prod_{t=1}^Tp(\mathbf{s}_{t+1}|\mathbf{s}_t,\mathbf{a}_t,\pmb{\rho}_t)p(\mathbf{a}_t)\prod_{t=2}^Tp(\pmb{\rho}_t|\mathbf{a}_t,\pmb{\rho}_{t-1})\nonumber
\end{equation}

As seen from Fig.~\ref{hmm03}, the transition between phases depends on the previous phase and the measured interaction $\mathbf{a}_t$. The hidden states also satisfy the Markov property, which allows efficient ways for parameter estimation.

\subsection{Model Selection}\label{MS}

For an HMM with $N$ phases, the model order $N$ is directly related with the fit of the model. However, an improvement in fit comes with a cost of a quadratic increase in number of model parameters. A criterion for model selection that solves the trade-off between the fit and model complexity is therefore needed. The number of hidden states can be selected with Bayesian information criterion (BIC) \cite{schwarz1978estimating}. BIC tries to find a model that optimally balances the trade-off between the model fit and the number of model parameters (i.e. the model complexity).

To estimate the model order for an HMM with normally distributed observation probabilities, the formulation of BIC is 

\begin{equation}
BIC = -2\log(L)+p\log(T)
\end{equation}

\noindent where $\log(L)$ is the log-likelihood measuring the fit of the model, $p=N^2+2N-1$ refers to the number of parameters in the model ($N-1$ for the initial distribution, $N(N-1)$ for the other), and $T$ is the number of observations. Model selection is based on the BIC value computed for different orders of the HMM. We choose the model with the lowest value. 

\subsection{Model Learning}\label{ML}

The impedance controller parameters for each individual phase are learned with the method presented in \cite{suomalainen2017}. For the model defined in Sect.~\ref{Model} we adapt the expectation-maximization (EM) algorithm from \cite{kroemer2014learning} to learn from multiple demonstrations the model parameters $\theta=(w, A, B, \Sigma)$. As shown in Fig.~\ref{em02}, the algorithm starts with an initial guess of the model parameters, $\theta_0$, in addition to the position $S$ and interaction $F$ recorded during a demonstration. The model parameters are initialized using k-means clustering. During the E-step, a forward-backward algorithm is used to compute the forward and backward messages. The marginal likelihood $\gamma_t(j)=p(\pmb{\rho}_t=j|\mathbf{a}_{1:N})$ and the joint distribution $\zeta_t(i,j)=p(\pmb{\rho}_t=i, \pmb{\rho}_{t+1}=j|\mathbf{a}_{1:N})$ of the hidden phases are computed from the forward and backward messages. The M-step uses as input $\gamma_t(j)$ and $\zeta_t(i,j)$ computed during the E-step and updates the model parameters $\hat{\theta}=(\hat{w}_j,\hat{A}_j,\hat{B}_j,\hat{\Sigma}_j)$ using the weighted linear regression for parameters related to state dynamics and weighted logistic regression for the parameter related to the probabilistic classifier. The algorithm iterates until either the estimation converges or the maximum number of iterations is reached.

\colorlet{colB1}{red!40}
\colorlet{colB2}{cyan!40}
\colorlet{colB3}{blue!40}
\colorlet{colB4}{black!100}
\colorlet{colBorder}{gray!70}
\tikzset
{mybox/.style=
	{rectangle,rounded corners,drop shadow,minimum height=2cm,
		minimum width=2cm,align=center,fill=#1,draw=colBorder,line width=3pt
	},
	myarrow/.style=
	{draw=#1,line width=5pt,-stealth,rounded corners
	},
	mylabel/.style={text=#1}
}

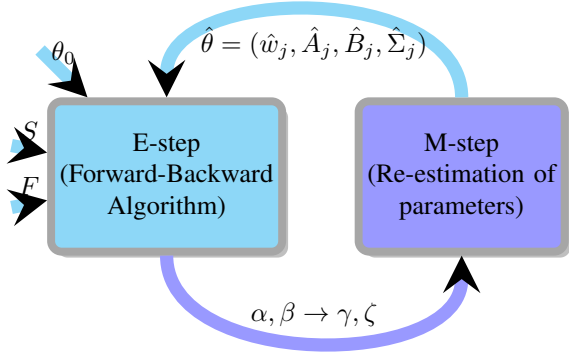
\begin{figure}[hbt]
	\centering
	\begin{tikzpicture}[scale=0.5]
	\node[mybox=colB2] (B2) at (-2,-3) {E-step\\(Forward-Backward\\Algorithm)};
	\node[state,draw=none] (d3) [left of=B2] at (-4,1) {};
	\node[state,draw=none] (d1) [left of=B2] at (-5,-2) {};
	\node[state,draw=none] (d2) [left of=B2] at (-5,-4) {};
	\node[mybox=colB3,right=of B2] (B3) at (1,-3) {M-step\\(Re-estimation of\\parameters)};
	\draw[myarrow=colB2] (d1) -- (B2);
	\draw[myarrow=colB2] (d2) -- (B2);
	\draw[myarrow=colB2] (d3) -- (B2);
	\draw[myarrow=colB2] (B3) edge[bend right=90] node[mylabel=colB4,below]{$\hat{\theta}=(\hat{w}_j,\hat{A}_j,\hat{B}_j,\hat{\Sigma}_j)$} (B2);
	\draw[myarrow=colB3] (B2) edge[bend left=-90] node[mylabel=colB4,above]{$\alpha,\beta\to\gamma,\zeta$} (B3);
	\path (d1) -- node[mylabel=colB4,above]{$S$} (B2);
	\path (d2) -- node[mylabel=colB4,above]{$F$} (B2);
	\path (d3) -- node[mylabel=colB4,above]{$\theta_0$} (B2);
	\end{tikzpicture}
	\caption{Block diagram representation of the EM algorithm}
	\label{em02}
\end{figure}

\subsubsection{Expectation-step (E-step)}

During the E-step, the marginal likelihood $p(\pmb{\rho}_t=j|\mathbf{a}_{1:N})$ and the joint distribution $p(\pmb{\rho}_t=i, \pmb{\rho}_t=j|\mathbf{a}_{1:N})$ of the hidden phases given the observed sequence of variables are computed separately for each demonstration using forward-backward message passing.
The intermediate variables obtained from running E-step separately for all $D$ demonstrations are conditional hidden phase probability $\gamma=\left[\begin{array}{cccc}\gamma^{(1)} & \gamma^{(2)} & \cdots & \gamma^{(D)} \end{array} \right]$ and joint hidden probability $\zeta=\left[\begin{array}{cccc}\zeta^{(1)} & \zeta^{(2)} & \cdots & \zeta^{(D)} \end{array} \right]$.

\subsubsection{Maximization-step (M-step)}

During the maximization step of the EM algorithm, the parameters of the HMM are computed for the maximization of expected log-likelihood of the observed and hidden variables. We define the total marginal likelihood as a concatenation of the marginal likelihoods of each demonstration as 

\begin{equation}
\gamma=\left[\begin{array}{cccc}
\gamma^{(1)} & \gamma^{(2)} & \cdots & \gamma^{(D)}
\end{array} \right]
\end{equation} 

The state and interaction matrices are estimated using weighted linear regression. Each column of matrix $X$ consists of the concatenated state $\mathbf{s}$ and interaction $\mathbf{a}$ of a demonstration followed by another demonstration until all $D$ demonstrations are included. Matrix $Y$ is next state matrix consisting of $T_1+T_2+\cdots+T_D$ columns corresponding to the a sampled next state of $D$ demonstrations. $T_1, T_2,\cdots,T_D$ represents length of each demonstrations. Formally, we write

\begin{equation}
X=\left[\begin{array}{ccccc}\mathbf{s}_1^{(1)} \cdots \mathbf{s}_{T_1}^{(1)} & \mathbf{s}_1^{(2)} \cdots \mathbf{s}_{T_2}^{(2)} \cdots \mathbf{s}_1^{(D)} \cdots \mathbf{s}_{T_D}^{(D)}\cr \mathbf{a}_1^{(1)} \cdots \mathbf{a}_{T_1}^{(1)} & \mathbf{a}_1^{(2)} \cdots \mathbf{a}_{T_2}^{(2)} \cdots \mathbf{a}_1^{(D)} \cdots \mathbf{a}_{T_D}^{(D)}\end{array}\right]
\end{equation}

where the superscript $(1),(2),\cdots,(D)$ is used to indicate the data is from demonstration $1,2,\cdots D$ ,respectively. Similarly, $Y$ is defined as

\begin{equation}
Y=\left[\begin{array}{ccccc}\mathbf{s}_2^{(1)} \cdots \mathbf{s}_{T_1}^{(1)} & \mathbf{s}_2^{(2)} \cdots \mathbf{s}_{T_2}^{(2)} \cdots \mathbf{s}_2^{(D)} \cdots \mathbf{s}_{T_D}^{(D)}\end{array}\right]
\end{equation}

The new estimates of the parameter matrix from the combined demonstrations can be derived from 
\begin{equation}
\left[\begin{array}{cc} A_j & B_j\end{array}\right]=\min\limits_{A_j,B_j}\norm{Y-\left(\begin{array}{cc} A_j & B_j\end{array}\right)X}^2
\end{equation}
which leads to 
\begin{equation}
\left[\begin{array}{cc} A_j & B_j\end{array}\right]=YW_jX^T\left(XW_jX^T\right)
\end{equation}

where the superscript $T$ indicates the transpose of a matrix and $W_j$ is the combined diagonal matrix corresponding to $\gamma$, where the $t^{th}$ diagonal element is given by $\left[W\right]_{tt}=p(\pmb{\rho}_t=j|\mathbf{a}_{1:N+1}, \theta_{old})$. 

Let the expected next state for each demonstration, given the updated $A$ and $B$ matrices, be represented by  $\mu_{ji}^{(1)}=A_j\mathbf{s}_i^{(1)}+B_j\mathbf{a}_i^{(1)},\mu_{ji}^{(2)}=A_j\mathbf{s}_i^{(2)}+B_j\mathbf{a}_i^{(2)}\cdots\mu_{ji}^{(D)}=A_j\mathbf{s}_i^{(D)}+B_j\mathbf{a}_i^{(D)}$. A variable $\Delta \mathbf{s}$ which represents the difference between the concatenated next state and the expected next state is given by

\begin{equation}
\Delta \mathbf{s}=\left[\begin{array}{cc}\left(\mathbf{s}_{t+1}^{(1)}-\mu_{ji}^{(1)}\right) & \left(\mathbf{s}_{t+1}^{(2)}-\mu_{ji}^{(2)}\right)\cdots\left(\mathbf{s}_{t+1}^{(D)}-\mu_{ji}^{(D)}\right)\end{array} \right]
\end{equation}
Using the combined $\Delta \mathbf{s}$, we calculate the new estimates of the covariance matrices as
\begin{equation}
\Sigma_j=\frac{\sum\limits_{i=1}^Np(\pmb{\rho}_i=j|\mathbf{a}_{1:N+1}, \theta_{old})(\Delta \mathbf{s})^T(\Delta \mathbf{s})}{\sum\limits_{k=1}^Np(\pmb{\rho}_k=j|\mathbf{a}_{1:N+1}, \theta_{old})}
\end{equation}

To estimate the phase transition probability parameter $w$, weighted logistic regression is used. Logistic regression is calculated iteratively using gradient descent. The gradient of the $j$-th phase given the $i$-th phase can be obtained by derivative of the phase transition distribution with respect to the phase transition probability parameter $w$. After derivation and simplification, the gradient for transition from $\pmb{\rho}_{t-1} = i$ to $\pmb{\rho}_{t} = j$ can be evaluated in terms of three matrices: $F, L$ and $P$. The columns of matrix $F$ contain the features of the sampled states $\phi(a_t)$, matrix $L$ is the joint phase probability containing weights from E-step ($[L]_{tj}=p(\pmb{\rho}_{t-1}= i,\mathbf{\pmb{\rho}}_t = j|\mathbf{z}_{1:N+1}, \theta_{old})$) and matrix $P$ contains the weights of the phase transition probabilities. The combined gradient of the weighted log-likelihood with respect to $w$ are given by

\begin{fleqn}
	\begin{align}
	G = F^{(1)}(P^{(1)}-L^{(1)})+F^{(2)}(P^{(2)}-L^{(2)})+\cdots \\ \nonumber
	+F^{(D)}(P^{(D)}-L^{(D)})
	\end{align}
\end{fleqn}
where the superscripts $1\dots D$ denote the demonstrations. Then the update of each weight matrix $w_{ij}$ containing weights of transitioning from $\pmb{\rho}=i$ to $\pmb{\rho}=j$ is done using gradient descent at each time step from the previous time step, where $k$ represents iteration time steps for gradient descent optimization
\begin{equation}
w_{k+1}= w_k-\lambda G 
\end{equation} 
After the M-step has been completed, the algorithm computes the E-step again with the new parameters, as visualized in Fig.~\ref{em02}. 

\subsection{Reproduction}
During online operation, the phase changes need to be detected to switch between impedance controllers with different parameters. For each phase the controller parameters are learned using \cite{suomalainen2017}. The current phase is estimated as the most likely one given the online observations and the learned model using the forward algorithm. In forward algorithm, the forward message $\pmb{\alpha}_t (j)$ of a phase is the joint probability of all observed data $\mathbf{z}_t=\{\mathbf{s}_t, \mathbf{a}_t\}$ up to time $t$. Therefore the probability of being in phase $\pmb{\rho}_t$ is given by 
\begin{equation}
	\pmb{\alpha}_t (j) = p(\mathbf{z}_{1:t+1}, \pmb{\rho}_t=j).
\end{equation}
The first forward message is initialized according to 
\begin{equation}
	\pmb{\alpha}_1(j) = p(\mathbf{z}_2|\pmb{\rho}_1,\mathbf{z}_1)p(\pmb{\rho}_1=j|\mathbf{z}_1)p(\mathbf{z}_1)
\end{equation} 
and the remaining messages are computed recursively using
\begin{equation}
	\pmb{\alpha}_t(j)=p(\mathbf{z}_{t+1}|\mathbf{\rho}_t=j, \mathbf{z}_t)\sum_i\pmb{\alpha}_j(t-1)p(\pmb{\rho}_t=j|\pmb{\rho}_{t-1}=i).
\end{equation}
After computing $\pmb{\alpha}_t(j)$, the current phase is estimated as
\begin{equation}
\rho = \arg\max_{j} \pmb{\alpha}(j).
\end{equation}
As the learned model includes hysteresis effects, they don't need to be specified manually.
\section{EXPERIMENTS AND RESULTS}
\label{EXPERIMENTS}
The first research question is how well we can learn the state dynamics model, i.e. the smoothing problem. The second question is to evaluate the prediction of phase transition during reproduction, i.e. the filtering problem. 

The proposed method was implemented and evaluated experimentally on a KUKA LWR4+ robotic arm with a six-axis ATI mini 45 F/T sensor rigidly mounted between the robot's flange and tool. A regular pen was attached to the F/T sensor for the valley setup and a hose-coupler for the hose-coupling setup. 
The robot was interfaced to an external computer through the KUKA \textit{Fast Research Interface} (FRI) protocol at 100 Hz \cite{schreiber2010fast}.
The control law for the Cartesian impedance control of the KUKA LWR4+ through FRI is  
\begin{equation}
	\tau=J^T\left(\text{diag}(\textbf{k}_{FRI})\left(\mathbf{x}^*-\mathbf{x}\right)+D(d_{FRI})\textbf{v}+\textbf{F}_{FRI}\right)+\mathbf{f}_{dyn}
\end{equation}
where $\mathbf{x}^*$ is the desired position, $\mathbf{x}$ the current position, $\text{diag}(\textbf{k}_{FRI})$ is a diagonal matrix constructed of the gain matrix $\textbf{k}_{FRI}$, $D(d_{FRI})$ a linear damping term and $\mathbf{f}_{dyn}$ the feed-forward dynamics of the robot. We implemented our controller through the $\textbf{F}_{FRI}$ by setting $\textbf{k}_{FRI}=0$ and $\textbf{F}_{FRI}=K(\pmb{x}^*-\pmb{x})$, getting a controller equal to (\ref{eqt:imp_control}) where $\mathbf{f}_{dyn}$ is managed by the KUKA's internal controller.

\begin{figure}[htb]
	\centering
	\begin{subfigure}{.25\textwidth}
		\centering
		\includegraphics[width=4cm,height=6cm,keepaspectratio]{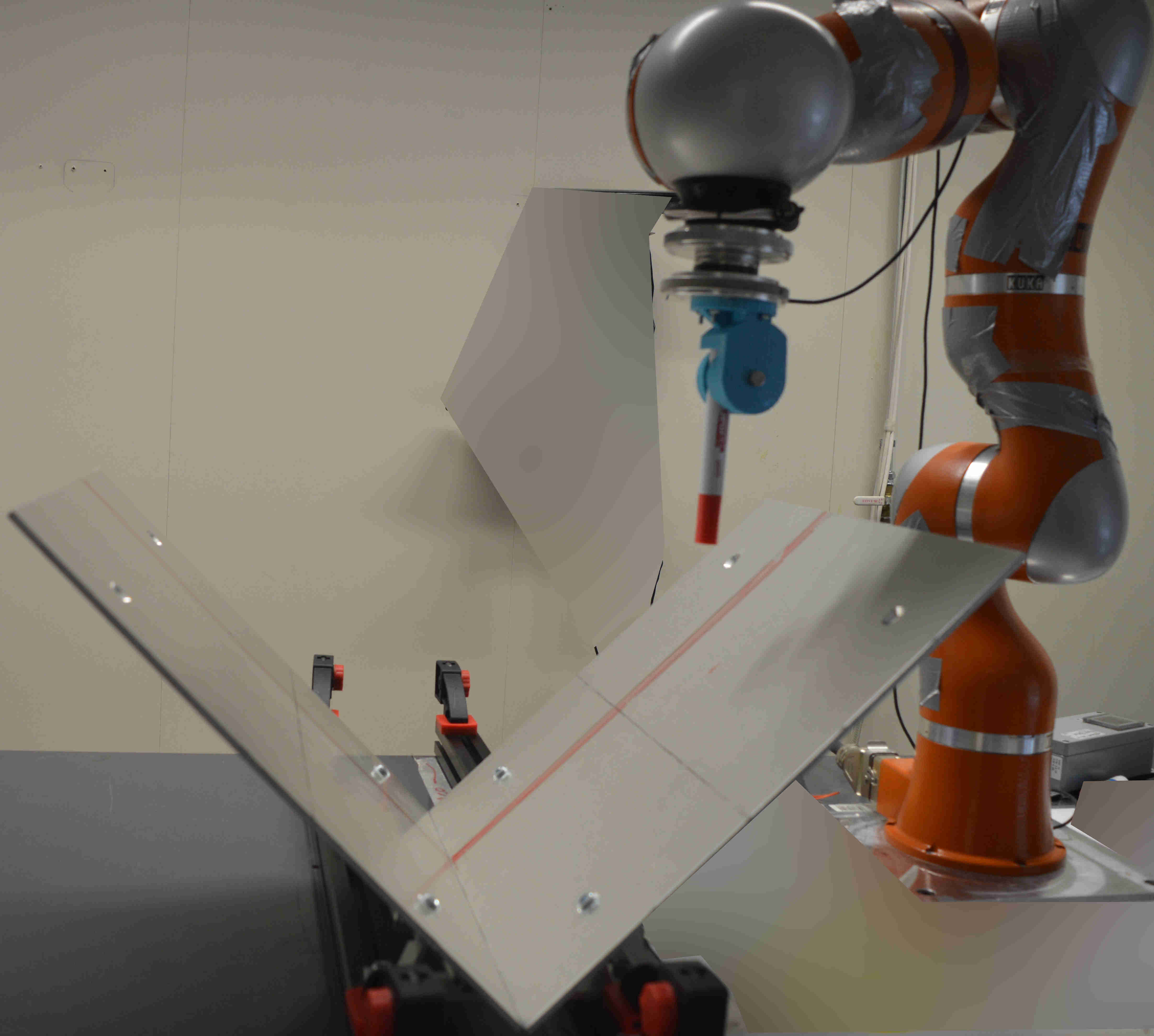}
		\caption{Experiment setup from SP 1}
		\label{setup21}
	\end{subfigure}%
	\begin{subfigure}{.25\textwidth}
		\centering
		\includegraphics[width=4cm,height=5cm,keepaspectratio]{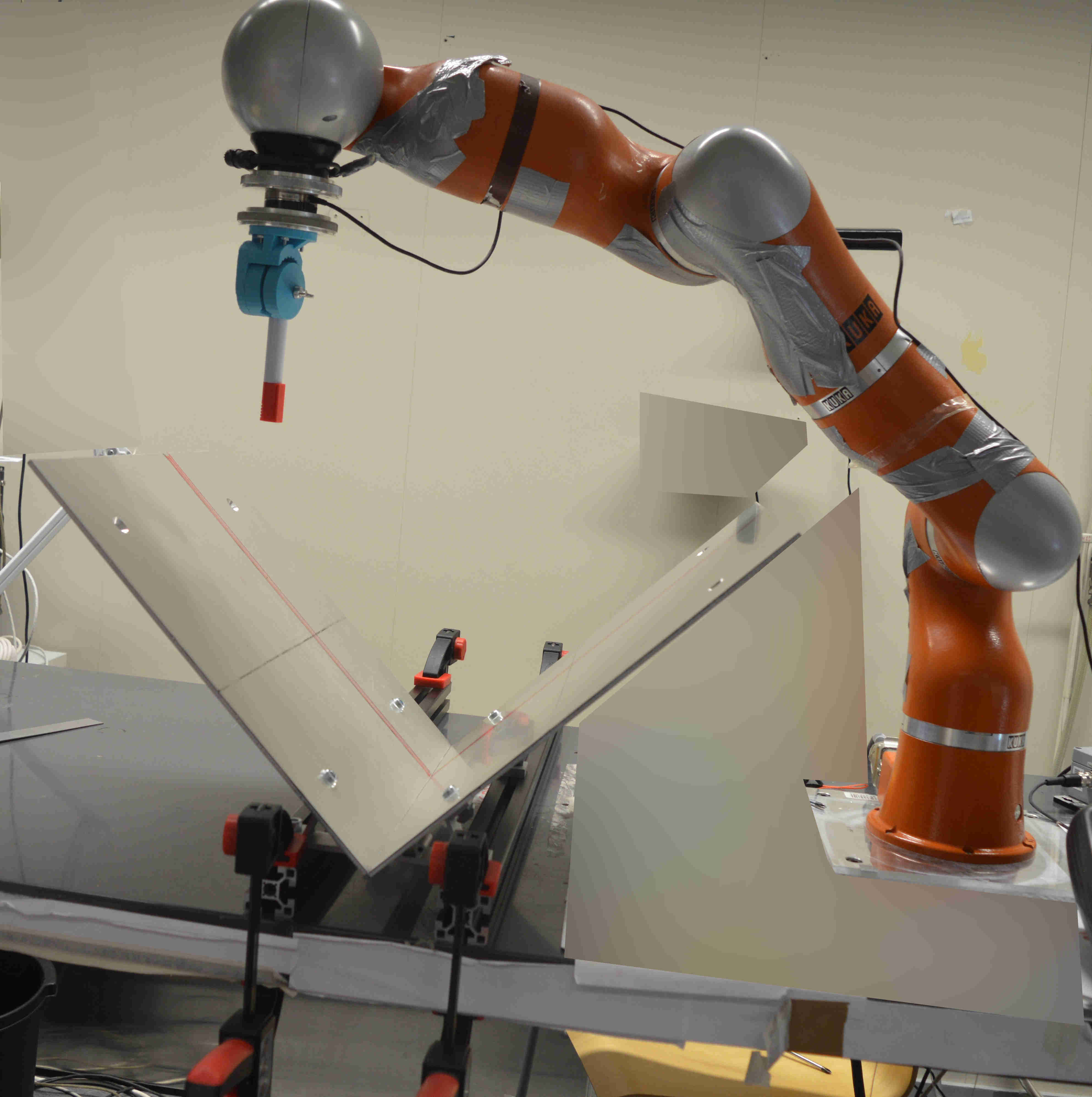}
		\caption{Experiment setup from SP 2}
		\label{setup22}
	\end{subfigure}
	\caption{Experiment setup for demonstrations from two different starting positions (SP)}
	\label{setup20}
\end{figure}

We carried out three experiments where a human instructor grabbed the flange of the robot and moved the robot through the trajectory. The first experiment was performed on a valley consisting of two aluminium plates set on 45 degrees angle on the table as shown in Fig.~\ref{setup20}. The demonstrated task included three phases. The first phase was an unconstrained motion in which the tool moves directly downwards. The second and third phases were the constrained motions; in the second one, the pen slid downwards along one of the aluminium plates, and in the third one the sliding direction was along the bottom of the valley (i.e. towards the camera in Fig.~\ref{setup20}). Position and force data was recorded throughout the demonstrations.

In the second experiment, we performed the same task except that this time we had two different starting positions, one from left and the other from the right side of the valley as shown in Fig.~\ref{setup21} and \ref{setup22}. In both cases, the pen slid down the corresponding side of the valley until it reached the bottom, and then it slid towards the camera along the bottom of the valley. The main goal of the second experiment was to find a single model that can reproduce the tool alignment regardless of the initial contact point between the tool and the valley (i.e. show that the algorithm learns to take advantage of chamfers while being robust to positional errors).

\begin{figure}[h!]
	\centering
	\includegraphics[height=4cm]{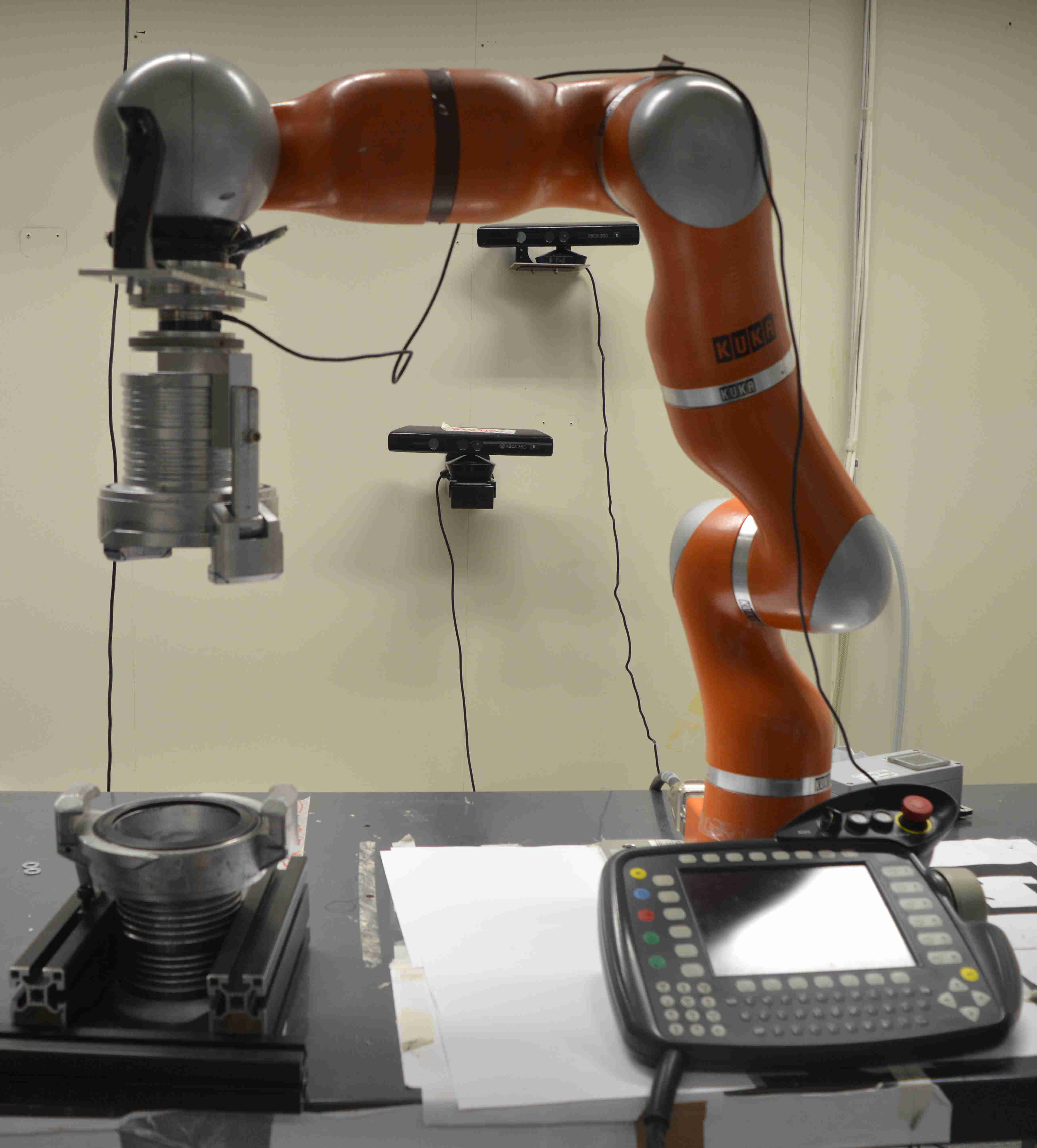}
	\caption{Hose-coupler Experimental setup}
	\label{setup31}
\end{figure}

As the final experiment, to make sure our method can also manage tasks that involve rotation, we use the hose-coupler experimental setup shown in Fig.~\ref{setup31}. The task includes two phases; the first phase is the unconstrained motion in which tool hose-coupler moves towards the other hose coupler attached to the table. The second phase rotates the tool, interlocking the hose couplers. For this task, pose and wrench data were recorded. 

\subsection{Model Learning}

The main objective was to segment a workpiece alignment task into phases and find model parameters that best fit each phase while learning the controller parameters with \cite{suomalainen2017}. Based on the obtained model parameters, we identified the phases during reproduction and activated the corresponding impedance control primitive.

In the valley setup experiments the position of the tool is considered the state in the HMM, as we assume that the orientation stays constant. 
Since this experiment does not require rotational motions, the feature vector is defined by the contact force concatenated with scalar one.

\begin{figure}[h!]
	\centering
	\begin{subfigure}{0.25\textwidth}
		\centering
		\includegraphics{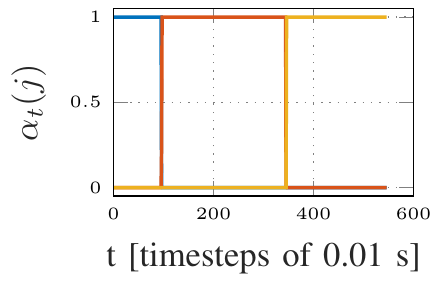}
		\caption{Estimated phases sequence}
		\label{est01}
	\end{subfigure}%
	\begin{subfigure}{0.25\textwidth}
		\includegraphics{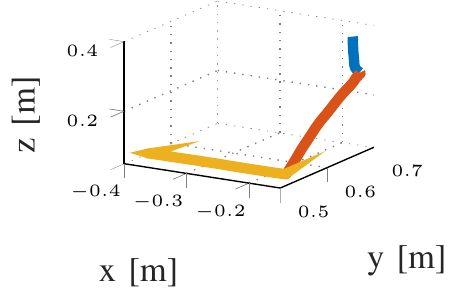}
		\caption{Position of the tool}
		\label{est02}
	\end{subfigure}
	\caption{Phases estimated by the normalized forward variable of HMM $\pmb{\alpha}_t(j)$ and corresponding tool position. The blue color corresponds to non-contact phase, the red and yellow colors corresponds to the sliding across the metal plate and sliding along the intersection of the two metal plates respectively.}
	\label{fig:test}
\end{figure}

To validate how well we can predict the state dynamics model, we first ran the EM algorithm for recorded data from a single demonstration to estimate the model parameters. The algorithm was able to predict the phases of the task as shown in Fig.~\ref{est01}. The marked 3D plot of the position of the tool corresponding to the estimated phase Fig.~\ref{est01} is shown in Fig.~\ref{est02}.  

\begin{figure}[h!]
	\centering
	\begin{subfigure}{0.25\textwidth}
		\centering
		\includegraphics{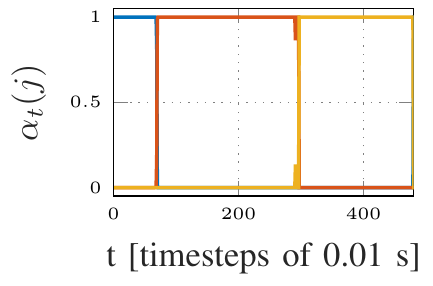}
		\caption{Phase sequence 1}
		\label{md03}
	\end{subfigure}%
	\begin{subfigure}{0.25\textwidth}
		\includegraphics{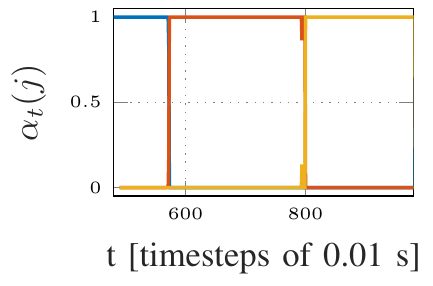}
		\caption{Phase sequence 2}
		\label{md04}
	\end{subfigure}
	\vskip\baselineskip
	\begin{subfigure}{0.5\textwidth}
		\centering
		\includegraphics{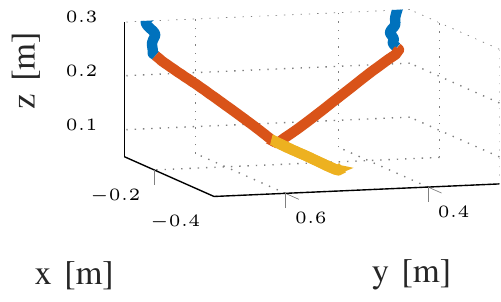}
		\caption{Position of tool}
		\label{md05}
	\end{subfigure}
	\caption{Phase sequence estimated by the normalized forward variable of HMM $\pmb{\alpha}_t(j)$ for two different demonstrations and their corresponding 3-d plot of tool position. The blue color corresponds to non-contact phase, the red and yellow colors corresponds to the sliding across the metal plate and sliding across the intersection of the two metal plates respectively.}
	\label{fig:test}
\end{figure}

To check if our implementation can learn that sliding across either metal plate is the same phase of the manipulation task, two demonstrations with different starting positions were used. We ran the EM algorithm described in Sect.~\ref{ML} to estimate the combined model parameters that generalize our model to both sides of our setup. As in the case of estimation from a single demonstration, the algorithm is able to estimate the phase sequence of the task as shown in the Fig.~\ref{md03} and \ref{md04} which correspond to the normalized forward variable of the HMM $\pmb{\alpha}$ for left and right side of the setup, respectively. The marked 3D plot of the position of the tool corresponding to the phase sequence of Fig.~\ref{md03} and \ref{md04} is shown in Fig.~\ref{md05}. 

\begin{figure}[h!]
	\centering
	\begin{subfigure}{0.25\textwidth}
		\centering
		\includegraphics{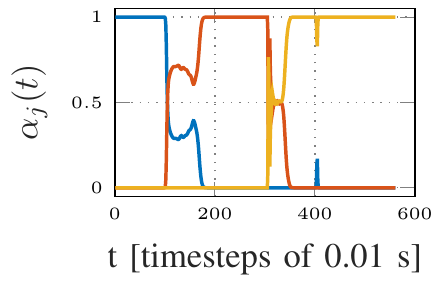}
		\caption{State-based estimated phases}
		\label{C01}
	\end{subfigure}%
	\begin{subfigure}{0.25\textwidth}
		\centering
		\includegraphics{h_est01.pdf}
		\caption{force-based estimated phases}
		\label{C02}
	\end{subfigure}%
	\caption{Estimated phase sequence based on state-based and interaction-based phase transitions}
	\label{cmp}
\end{figure}

In addition, we compared our approach in the valley setup with the state-based phase transition presented in \cite{kroemer2014learning} and show the results in Fig. \ref{cmp}. For the state-based transition, we took the relative position of the tool to a target position as the feature vector. A smoother phase transition is achieved when forces are considered as feature vector in Fig. \ref{C01} compared to the state based transition in Fig. \ref{C02}. Therefore, for compliant motions, the force as a feature vector can distinguish phases of an alignment task better compared to state-based phase transition.

To further show that the contact force provides better prediction of next states and validate the model of (\ref{eqt:statedynamics}), the model parameters estimated for each phase were used to predict the whole state dynamics. The average error variance of prediction with respect to the actual position data was calculated. The average error variance of each phase is given by 

\begin{equation}
E_{j}(\theta_i)=\frac{\sum\limits_{t=0}^{T-1}|\mathbf{s}_{t+1}(\theta_i)-\mathbf{s}_{t+1}|^2}{T-1}
\end{equation}
where $E_{j}(\theta_i)$ is average error variance corresponding to each phase $j$ using model parameters $\theta_i$. $\mathbf{s}_{t+1}(\theta_i)$ state prediction and $\mathbf{s}_{t+1}$ is the actual state data. 

\begin{figure}[htb]
	\centering
	\includegraphics{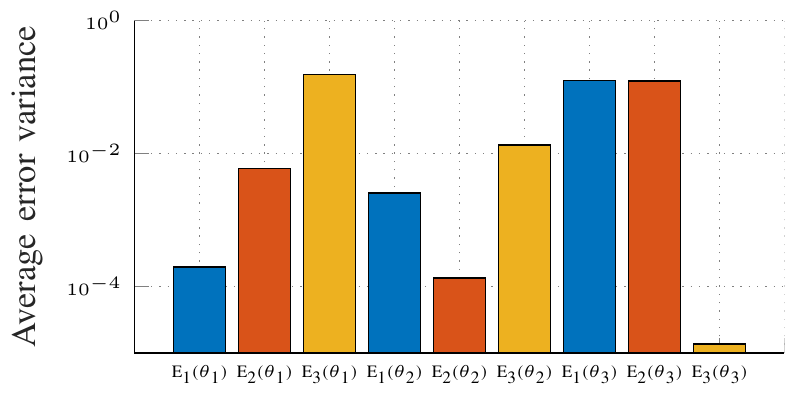}
	\caption{State dynamics prediction average error variance}
	\label{PE01}
\end{figure}

In Fig.~\ref{PE01}, the first three bars correspond to the first phase model parameters, the next three correspond to the second phase model parameters and the last three correspond to the last phase model parameters. The prediction accuracy of each phase is shown by a lower average error variance as indicated in the figure by $E_{1}(\theta_1)$, $E_{2}(\theta_2)$ and $E_{3}(\theta_3)$ respectively. To see the effect of feature vector variation on prediction of state dynamics, the average error variance of prediction of each phase dynamics using its own model parameters are compared i.e. $E_{1}(\theta_1)$, $E_{2}(\theta_2)$ and $E_{3}(\theta_3)$. $E_{3}(\theta_3)$ shows lower error, which indicates that an increased value in the feature vector (i.e. more interactions with enviroment) results in better prediction of states. This shows that taking force as feature vector will also improve the dynamics prediction for a task consisting of compliant motions.       

\begin{figure}[htb]
	\centering
	\includegraphics{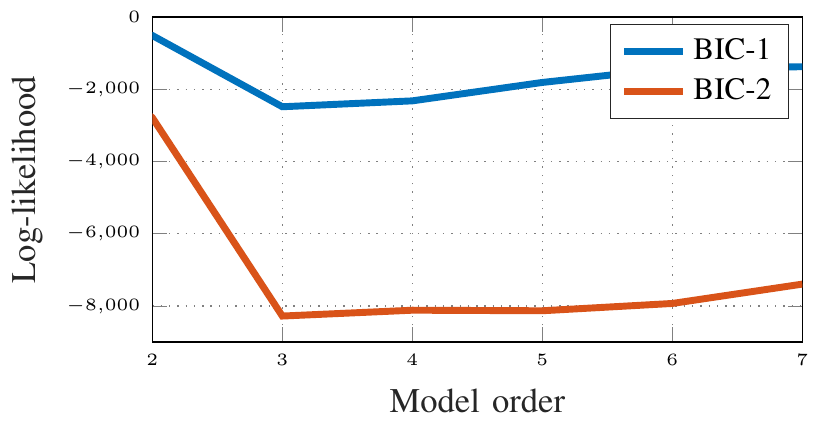}
	\caption{HMM order estimation for single (red) and two demonstrations (blue)}
	\label{aic01}
\end{figure}

BIC described in Sect.~\ref{MS} can be used to predict the model order for one or more demonstrations. In Fig.~\ref{aic01}, BIC for both demonstrations in the valley setup are plotted against the number of states $N$ of the HMM. According to BIC, the model with three states is the most appropriate model number for the demonstrated task, as seen from the lower log-likelihood. The estimated model order from single and from two demonstrations is labelled by BIC-1 and BIC-2, respectively. We note that from noisy demonstrations it was possible that more than 3 phases were detected. This, however, did not hinder the overall reproduction performance of our method, but simply required a new set of parameters to be learned. 

Similar approach was used to learn the model parameters of the hose-coupler from recorded demonstration data (pose and wrench). The model order was obtained using BIC and model parameters were estimated with the EM algorithm.
\subsection{Reproduction}
\begin{figure}[h!]
	\centering
	\begin{subfigure}{0.25\textwidth}
		\centering
		\includegraphics{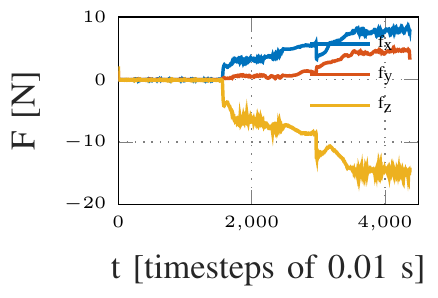}
		\caption{Contact Force}
		\label{PT01}
	\end{subfigure}%
	\begin{subfigure}{0.25\textwidth}
		\includegraphics{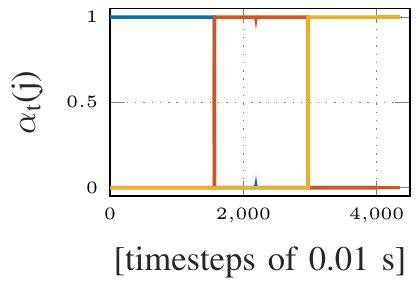}
		\caption{Phases sequence}
		\label{PT02}
	\end{subfigure}
	\vskip\baselineskip
	\begin{subfigure}{0.5\textwidth}
		\centering
		\includegraphics{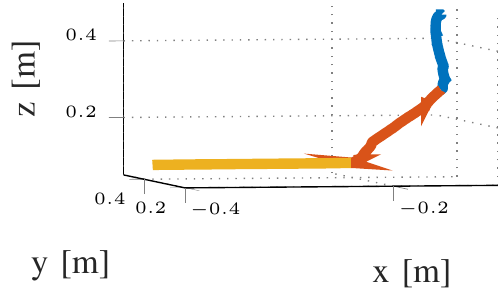}
		\caption{Tool Position}
		\label{PT03}
	\end{subfigure}
	\caption{Feature vector, corresponding phase sequence estimated by the normalized forward variable of HMM $\pmb{\alpha}_t(j)$ and 3-d plot of tool position during the reproduction of the task. The blue color corresponds to non-contact phase, the red and yellow colors corresponds to the sliding across the metal plate and sliding across the intersection of the two metal plates respectively.}
	\label{fig:hosereprod}
\end{figure}

The robot was able to reproduce the task using the learned model parameters from the valley and hose-coupler setups. For the hose-coupler setup, the feature vector and the phase sequence detected during reproduction are plotted side by side in Fig. \ref{fig:hosereprod}. As shown in Fig.~\ref{PT01}, the non-contact phase is characterized by zero contact force. During contact the $x$ and $z$ components of the contact force become non-zero as shown in Fig.~\ref{PT01} which results to phase transition indicated by quick change of the normalized forward variable seen in Fig. \ref{PT02}. During the last phase, contact force occurs also in the $y$-direction, resulting in the last transition. The trajectory with phases coloured respectively to Fig. \ref{PT02} is also given in Fig.~\ref{PT03}. 

\begin{figure}[h!]
	\centering
	\begin{subfigure}{0.25\textwidth}
		\centering
		\includegraphics{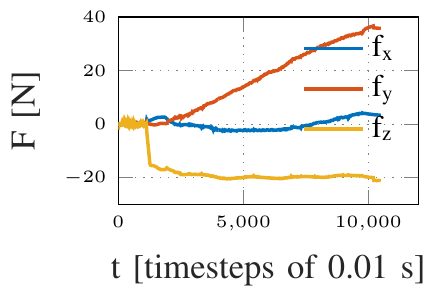}
		\caption{Contact Force}
		\label{HC01}
	\end{subfigure}%
	\begin{subfigure}{0.25\textwidth}
		\includegraphics{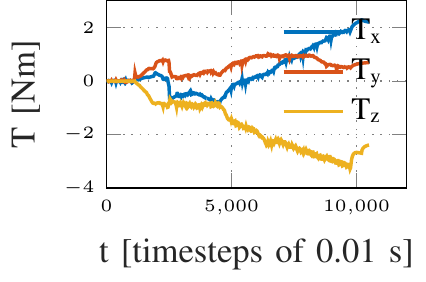}
		\caption{Contact Torque}
		\label{HC02}
	\end{subfigure}
	\vskip\baselineskip
	\begin{subfigure}{0.25\textwidth}
		\centering
		\includegraphics{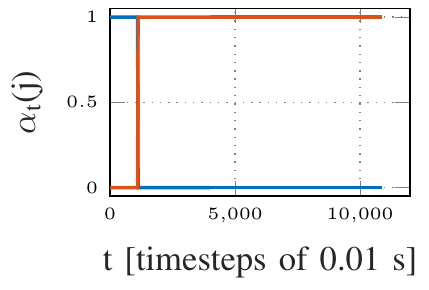}
		\caption{Phase sequence}
		\label{HC03}
	\end{subfigure}%
	\begin{subfigure}{0.25\textwidth}
		\centering
		\includegraphics{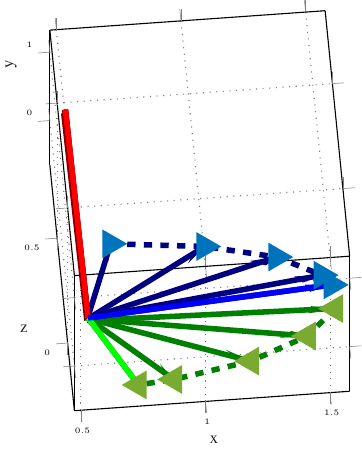}
		\caption{Rotation phase}
		\label{HC04}
	\end{subfigure}%
	\caption{Data recorded during reproduction of the hose-coupler experiment: contact force and torque, corresponding phase sequence (the blue color corresponds to non-contact phase, the red color corresponds to the rotation phase) and a 3-D plot of tool orientation during the reproduction of the final phase (rotation about z-axis)}
	\label{HC}
	\vspace{-0.6cm}
\end{figure}

The hose-coupler task was also successfully reproduced with the learned parameters. The feature vector (force and torque), forward variable $\pmb{\alpha}_t(j)$ and the orientation during the rotation phase are shown in Fig. \ref{HC}. As expected and seen in Fig.~\ref{HC01} and \ref{HC02}, during the non-contact phase the wrench is 0. As soon as contact occurs, the feature vector becomes non-zero and phase transition is detected, as seen in Fig.~\ref{HC03}. To show the rotation phase of the task, 3D position and orientation of the tool were plotted. As shown in Fig.~\ref{HC04}, during reproduction of the second phase rotation about the z-axis is achieved. We conclude that our method can successfully learn to reproduce a demonstrated assembly task consisting of phases with linear dynamics while taking advantage of chamfers to mitigate positional uncertainty.

\section{CONCLUSION}
\label{CONCLUSION}
In this paper, we presented a method that can learn to reproduce a human demonstration including in-contact compliant motions. First the method segments one or more demonstrations of the same task into phases comprised of motions with linear dynamics including compliance to facilitate in-contact motions. Then impedance controller parameters are learned individually for each phase using the method from \cite{suomalainen2017}. Finally, during reproduction our method sequences the learned impedance controllers to successfully reproduce the whole task. 

The phases were represented as hidden variables of an HMM. The model parameters can be learned from multiple demonstrations using the EM algorithm. The proposed model incorporates the observed interaction variables when predicting the transitions between the hidden phases. The framework is adapted to cope with non-constrained and constrained motion, handling hidden phase transition in both cases. Experiments evaluate the performance of the approach, showing that the off-line estimated model parameters are able to detect the current phase and transition between phases. The results show that for a task consisting of compliant motions the measured interaction-based phase transition allows the robot to predict the phase changes more accurately than state-based phase transition, resulting in better predictions overall. Our implemented algorithm can also be used for a task involving rotation as shown with the hose coupler experimental, demonstrating the ability to have different kind of interactions as the feature vector. Our method learns from the human to take advantage of possible chamfers in the task and due to compliance is robust towards positional errors and small changes in the environment. For tasks requiring nonlinear motions in free space or exact positioning in free space, it will still be necessary to use the pose in the feture vector such as in \cite{kroemer2014learning}.

The method presented in this paper can mitigate positional uncertainty by taking advantage of chamfers or other physical gradients. However, when such a gradient doesn't exist, our current method doesn't have the ability to recover from positional errors. When a human is faced with such a situation without vision, he will probe the environment in a manner possibly remembered from past experience. The next step in our research will be to see if this kind of search pattern could be learned from a human demonstration. This kind of ability to efficiently learn exception strategies from a human would enhance the method in this paper and further ease the use of robots in assembly tasks.





\bibliography{biblio}{}
\bibliographystyle{ieeetr}

\end{document}